\documentclass[letterpaper]{article} % DO NOT CHANGE THIS
\usepackage{aaai24}
\usepackage{times}  % DO NOT CHANGE THIS
\usepackage{helvet}  % DO NOT CHANGE THIS
\usepackage{courier}  % DO NOT CHANGE THIS
\usepackage[hyphens]{url}  % DO NOT CHANGE THIS
\usepackage{graphicx} % DO NOT CHANGE THIS
\usepackage{natbib}  % DO NOT CHANGE THIS AND DO NOT ADD ANY OPTIONS TO IT
\usepackage{caption} % DO NOT CHANGE THIS AND DO NOT ADD ANY OPTIONS TO IT
\usepackage{algorithm}
\usepackage{algorithmic}
\usepackage{multirow}
\usepackage{xcolor}

\usepackage{booktabs}

\usepackage{amsmath}
\usepackage{amssymb}

\usepackage{newfloat}
\usepackage{listings}
\DeclareCaptionStyle{ruled}{labelfont=normalfont,labelsep=colon,strut=off} % DO NOT CHANGE THIS
\floatstyle{ruled}
\newfloat{listing}{tb}{lst}{}
\floatname{listing}{Listing}
\title{PoetryDiffusion: Towards Jointly Semantic and Metrical Manipulation \\in Poetry Generation}
\author{
    Zhiyuan Hu\textsuperscript{\rm 1}\thanks{~~Equal Contribution},
    Chumin Liu\textsuperscript{\rm 2}\footnotemark[1],
    Yue Feng\textsuperscript{\rm 3},
    Anh Tuan Luu\textsuperscript{\rm 2},
    Bryan Hooi\textsuperscript{\rm 1}
}
\affiliations{
    \textsuperscript{\rm 1}National University of Singapore \quad
    \textsuperscript{\rm 2} Nanyang Technological University \quad
    \textsuperscript{\rm 3} University College London \\
    zhiyuan\_hu@u.nus.edu, chorlinglau@gmail.com, yue.feng.20@ucl.ac.uk \\anhtuan.luu@ntu.edu.sg, bhooi@comp.nus.edu.sg

}

\usepackage{bibentry}
\begin{document}

\maketitle

\begin{abstract}

Controllable text generation is a challenging and meaningful field in natural language generation (NLG). Especially, poetry generation is a typical one with well-defined and strict conditions for text generation which is an ideal playground for the assessment of current methodologies. While prior works succeeded in controlling either semantic or metrical aspects of poetry generation, simultaneously addressing both remains a challenge. In this paper, we pioneer the use of the Diffusion model for generating sonnets and Chinese SongCi poetry to tackle such challenges. 
%For semantic, Different from autoregressive generation and large language model (LLM), our PoetryDiffusion model, based on the Diffusion model, generates the complete sentence or poetry by taking into account the whole sentence information, resulting in improved semantic expression. 
In terms of semantics, our PoetryDiffusion model, built upon the Diffusion model, generates entire sentences or poetry by comprehensively considering the entirety of sentence information. This approach enhances semantic expression, distinguishing it from autoregressive and large language models (LLMs).
For metrical control, its constraint control module which can be trained individually enables us to flexibly incorporate a novel metrical controller to manipulate and evaluate metrics (format and rhythm).
The denoising process in PoetryDiffusion allows for the gradual enhancement of semantics and flexible integration of the metrical controller which can calculate and impose penalties on states that stray significantly from the target control distribution. Experimental results on two datasets demonstrate that our model outperforms existing models in terms of automatic evaluation of semantic, metrical, and overall performance as well as human evaluation.
Codes are released to \url{https://github.com/ChorlingLau/PoetryDiffusion/}.

\end{abstract}

\section{Introduction}

%The development of deep learning has made a significant impact on natural language generation (NLG). Seq2Seq model \cite{sutskever2014sequence} first led to breakthroughs in the deep learning era. GAN \cite{goodfellow2020generative}, VAE \cite{kingma2013auto} and pre-trained language models, including BART\cite{lewis2019bart}, T5 \cite{2020t5}, and GPT3 \cite{brown2020language}, etc., led to major advances in NLG. These have achieved state-of-the-art performance on many NLG tasks.
Deep learning has greatly influenced natural language generation (NLG). Models like Seq2Seq \cite{sutskever2014sequence}, GAN \cite{goodfellow2020generative}, VAE \cite{kingma2013auto}, pre-trained language models, and LLMs have led NLG advancements.
Among these, controllable text generation (CTG) is an emerging area within NLG and it is important and practical to consider specific constraints. Poetry generation stands out as a distinct domain with its unique characteristics, demanding not just coherent semantics but also strict adherence to metrical rules tied to format and pronunciation. These intricate requirements present the dual challenge of mastering semantics and metrics in tandem. In comparison to other coarse-grained CTG tasks like attribute-based generation (involving topics, emotions, and keywords), dialogue generation, and storytelling, poetry generation holds a unique position. Particularly evident in forms like sonnets and Songci, it necessitates adherence to well-defined and demanding metrics. Such specificity makes metrical poetry an ideal testing ground to validate the potency of the latest methodologies. Moreover, the available data resource for poetry is of unparalleled quality within the NLP domain, laying a robust foundation for our ensuing endeavors.

\begin{figure}
%\vspace{-\topsep}
    \centering
    \includegraphics[width=0.9\linewidth]{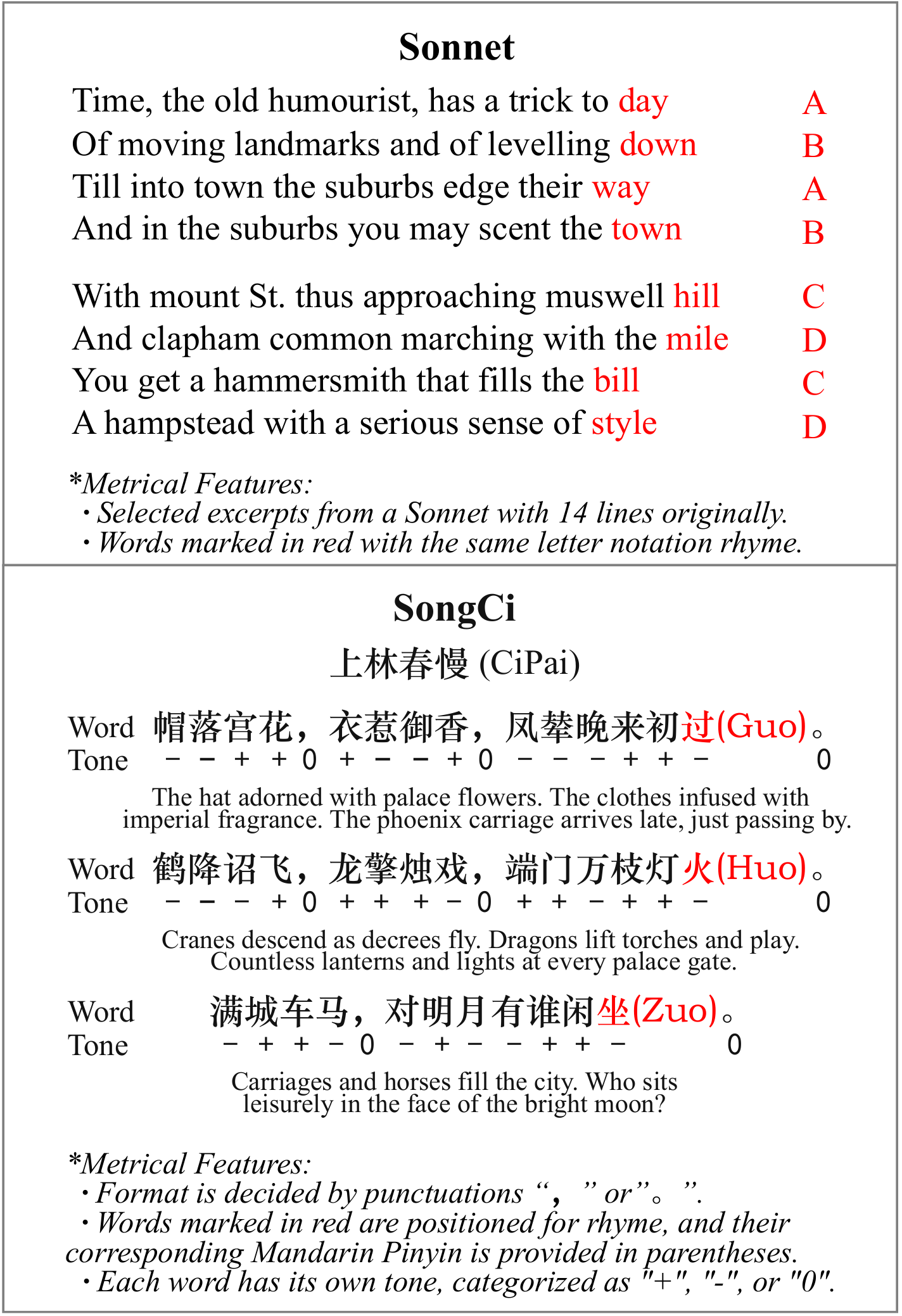}
    %\vspace{-\topsep}
    \caption{Examples of Sonnet and SongCi}
    %\vspace{-7mm}
    \label{example}
\end{figure}

Sonnet and SongCi are two classical and famous forms of poetry, which share two major characteristics: (1) The poems must adhere to special \textbf{format restrictions}. Sonnets must have 14 lines; similarly, the number of lines as well as the length of each line in SongCi is prescribed by the corresponding CiPai (Title of SongCi) (2) The chosen words must be consistent with specific \textbf{rhythm rules}. The last word of each line in Sonnets, as shown in Figure~\ref{example}, should follow the rhyme scheme "ABAB CDCD EFEF GG". In SongCi, the rhyme rule of the last word in every line is also set by its CiPai. In this example, the phonetic transcription of each word is "Guo", "Huo" and "Zuo", corresponding to one of 16 rhyme rules, "o". In addition, every word in SongCi must comply with the tone rule (Ping, Ze), which dictates pronunciation requirements. Level and oblique tones can be symbolized as "+", "-", and "0" (without a tone requirement). %\zhiyuan{to be detailed}

In previous CTG tasks, GAN and VAE models have been widely adopted as popular frameworks. 
%They are also widely employed in poetry generation. 
However, \citet{9555209} points out their limitations. Several key challenges associated with GANs encompass slow convergence, instability, vanishing gradients, mode collapse, and catastrophic forgetting. VAEs also suffer from posterior collapse, where the model ignores the latent variable and generates less diverse samples. Moreover, even the powerful LLMs, especially ChatGPT\cite{openai2021gpt35turbo}, surprise us with excellent generation capability, yet adhering strictly to specific instructions remains a challenge. \citet{pu2023chatgpt} conduct empirical studies that demonstrate ChatGPT's superiority over some previous SOTA models according to automated metrics. Despite this, notable discrepancies persist between ChatGPT's output and human-authored content. Our experiments also highlight this issue, revealing that ChatGPT's BLEU, ROUGE, and other semantic scores, which gauge the quality of generated poetry, fall short of our proposed method's scores. Additionally, the performance on metrics employed to evaluate ChatGPT's ability to adhere to metrical instructions and generate accurate metrical structures exhibits subpar performance, particularly in the SongCi dataset. It is worth noting that LLMs are trained on extensive corpora and derive their capabilities from instruction tuning. However, their generalization extent is uncertain, and further instruction tuning is resource-intensive and might compromise their original text generation quality. Although we might consider adopting efficient parameter-tuning techniques, such as Lora \cite{hu2021lora}, to mitigate this challenge, fine-tuning LLMs, especially those Billion-level models, remain more complex than fine-tuning our model, which has only 87 Million parameters.

%\cite{pu2023chatgpt} conducted some empirical findings showing that ChatGPT outperforms some previous SOTA models on automatic metrics, but there are substantial disparities between its generated texts and human-written texts. This problem is also observed in our experiments that the BLEU and ROUGE score of ChatGPT to measure the generated poetry quality is lower than our proposed method. And the metrics performances which assess whether ChatGPT can follow our metrical instruction and generate the correct metrical structure also underperform, especially in the SongCi dataset. LLM is trained on the general corpus and its ability comes from instruction tuning. The generalization can not be guaranteed and additional instruction tuning is highly resource-consuming and possibly harmful for its previous generation capability. Even if we can adopt the efficient parameter-tuning method like Lora \cite{} to alleviate this problem,  it is still more challenging to finetune LLMs(at least Billion level) rather than tuning our model only with 87 Million parameters.

%Additionally, VAEs can struggle with modeling complex distributions, and the choice of the prior distribution can affect the quality of generated samples.

Except for the aforementioned drawbacks about semantic performance and generating correct metrics based on the instructions, most previous poetry generation works \cite{zhang2014chinese,ghazvininejad2016generating,benhardt2018shall,van2020automatic,tian2022zero}, solely concentrated on modeling the semantics, do not explicitly enforce metrical constraints despite evaluating the metrical performance. Only two works SongNet\cite{li2020rigid} and MRCG\cite{zhang2019generating} directly incorporate the metrical rules representation into the generative model. However, both works utilize similar methods of encoding metrical rules into continuous representations and concatenating them with word embeddings, making it difficult to achieve satisfactory semantic performance when metrical features are combined in the modeling phase.

%\zhiyuan{elaborate the challenge of this task and the drawback of other models}

To address the challenges mentioned above, we propose the PoetryDiffusion model, which combines a Diffusion model for semantics with a metrical controller for metrics. Unlike other generative models, our diffusion approach introduces controlled noise through diffusion steps and then learns the reverse process to generate desired data. This design enhances training stability and generation quality. Moreover, this generative module and constraint separation increases adaptability for different generation tasks.

Specifically, the Diffusion model utilizes a noising process to transfer poetry representation into a normal distribution and samples from it as the input for the denoising phase, which reduces the noise and reverts it to the original poetry. The noising process is similar to gradually "masking" tokens, phrases, or certain dimensions of the representation. On the other hand, the denoising phase aims to "predict" the masked information and evaluate the success rate in each step. This mechanism ensures that the model captures all information of poetry rather than continuing to predict words based on wrong words in an autoregressive generation. 
%The metrical controller, utilizing classifier guidance which is significantly precise and stable compared with other generative models especially LLMs, effectively encodes metrical rules into a representation and evaluates the validity of the rules encoding. 
The metrical controller employs classifier guidance, which offers notably higher precision and stability than other generative models, particularly LLMs. This approach adeptly incorporates metrical rules into a representation while also assessing the validity of the encoded rules.
This allows for individual training and flexible integration, enabling efficient manipulation and assessment of metrics. Furthermore, when combining these two components to generate poetry, the modules for each step are updated based on feedback from Diffusion and controller in the previous step which indicates the accuracy of prediction for masked semantics and metrics.

To summarize, our contributions are as follows:
\begin{itemize}
% %\vspace{-\topsep}
  \setlength{\itemsep}{1pt}
  \setlength{\parskip}{1pt}
  \setlength{\parsep}{1pt}
    \item We propose the PoetryDiffusion model, which employs the Diffusion model to optimize the poetry semantic performance, leverages the metrical controller to model the metrical rules, and combines them flexibly and effectively, for the first time in poetry generation. 
    \item Comprehensive experiments through automatic semantic tests, metrical evaluations, case studies, and human evaluation on Sonnet and SongCi datasets demonstrate the effectiveness of our model. 
    \item The visualization and analysis of the stepwise process reveals how the PoetryDiffusion model integrates the semantics and metrics gradually.

\end{itemize}

%\zhiyuan{emphasize the contributions that we still address the metrical encoding question}

\begin{figure*}[htb]
%\vspace{-\topsep}
\centering
    \includegraphics[width=1\textwidth]{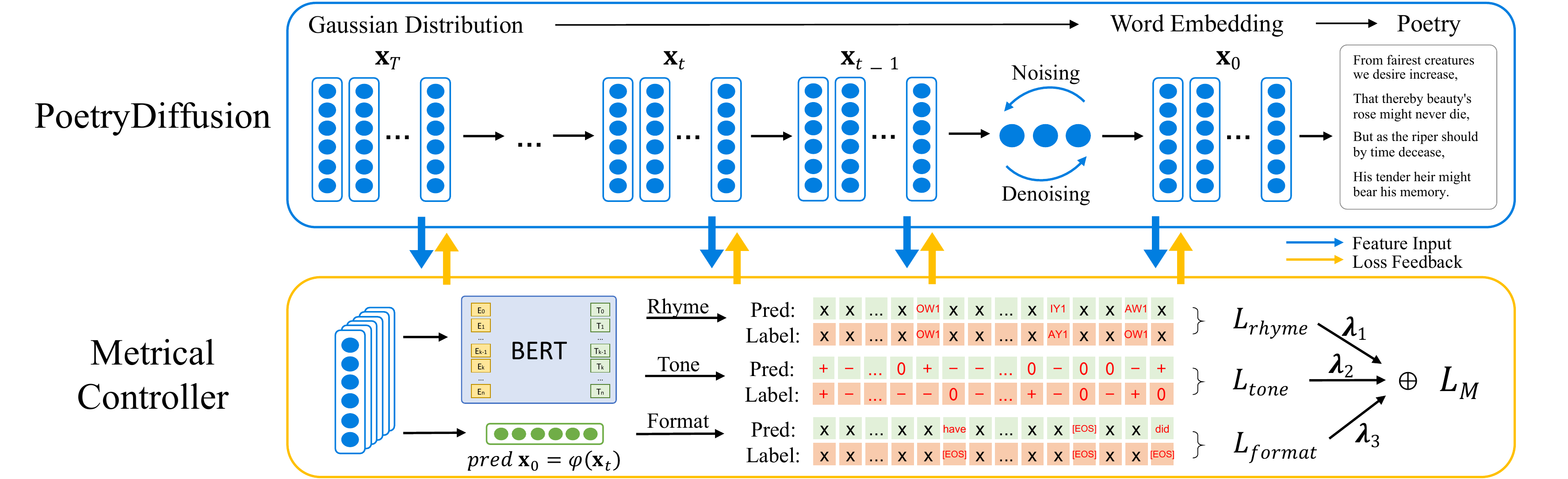}
    %\vspace{-\topsep}
    \caption{Model Architecture. PoetryDiffusion denoises ${\rm\bf x}_T$ to poetry ${\rm\bf w}$ based on joint loss ${\cal L}$ of each step.}
    %\vspace{-5mm}
    \label{Model}
\end{figure*}

\section{Related Work}

Controllable text generation refers to the task of generating text according to the given controlled element. \citet{hu2017toward} used differentiable approximation to discrete text samples, explicit constraints on independent attribute controls, and efficient collaborative learning of generators and discriminators to generate realistic sentences with desired attributes. \citet{betti-etal-2020-controlled} introduced Controlled TExt generation Relational Memory GAN which utilizes an external input to influence the coherence of sentence generation. Furthermore, \citet{li2022diffusion} proposed the Diffusion-LM to achieve several fine-grained controls. \citet{chen-yang-2023-controllable} incorporated different levels of conversation structures via Diffusion models to directly edit the prototype conversations.

Additionally, researchers also conducted some exploration based on the pretraining language model and large language model. \citet{zhang2022discup} introduced a method incorporating attribute knowledge into control prompts to steer a frozen casual language model to produce attribute-specific texts. \citet{sheng2021songmass} utilized the masked sequence to sequence pre-training and attention-based alignment modeling for lyric-to-melody and melody-to-lyric generation. \citet{zhang2023summit} employs multiple LLM as different roles in text generation to iteratively refine the generation results. \citet{zhou2023controlled} conducted extra instruction tuning for lexical, syntax, semantic, style, and length constraints based on the LLM.

In terms of poetry generation, \citet{yu2017seqgan} proposed SeqGAN, \citet{lin2017adversarial} introduced RankGAN, and \citet{che1702maximum} came up with MaliGAN for poem generation. \cite{chen2019sentiment} proposed the semi-supervised VAE model for sentiment control in poetry generation. \citet{yi2020mixpoet} leveraged the MixPoet to enhance the diversity and quality of the poem. \citet{deng2020iterative} utilized a Quality-Aware Masked Language Model to polish the draft poetry generated by the encoder-decoder model.

\section{Methodology}

\subsection{Overview}

% Generating only 1 token at a time makes it harder for them to conform to strict structural forms, as such forms may require longer-term global context from multiple lines, or going back to revise earlier written content like a human may do. In contrast, diffusion models rely on iterative refinement of the whole sequence and thus can utilize global context more easily, and avoid getting restricted based on earlier generated tokens.

As shown in Figure~\ref{Model}, the proposed method is divided into two parts. PoetryDiffusion is a Diffusion based framework. It converts poetry input into continuous word representation, then encodes it as a Gaussian distribution by noising. The denoising step samples an initial state from Gaussian distribution and reverts it into poetry. Metrical Controller evaluates metrics and transmits the loss to denoising steps, guiding the poetry to approach the control objectives.

\subsection{Diffusion Based Framework}
\label{PoetryDiffusion}
\textbf{Intuition} \indent As mentioned earlier, poetry is a well-structured literary form that demands thematic consistency and clarity, emphasizing coherence between its sub-sentences. Autoregressive models, which generate text word-by-word, have limitations that may lead to the accumulation of errors, resulting in off-topic or thematically inconsistent poetry. 
Additionally, generating only one token at a time makes it harder to conform to strict structural forms, which may require longer-term global context from multiple lines, or going back to revise earlier written content like a human would do. 
Therefore, considering the intricate control of poetic rhythm and the high demands placed on it, we opt for the Diffusion model \cite{sohl2015deep} as the semantic framework for generating poetry. It allows for comprehensive consideration of information from the entire poem during each iterative generation step and provides ample manipulative space for text controlling, especially format and rhyme scheme in poetry, throughout the iterations, and avoids getting restricted based on earlier generated tokens.%\chumin{Motivation for PoetryDiffusion}

To formulate the Diffusion model's principle, we define $q(\cdot)$ as its forward propagation distribution, while $p_\theta(\cdot)$ is the trainable backward one.  

Given a poem with $N$ words, we represent it as a sequence ${\rm\bf w}$ ($w_1, w_2, ... , w_N$), where $w_i$ stands for the $i$-th word of the poem. We adopt the methodology introduced by \cite{li2022diffusion}, wherein we tailor the continuous Diffusion model to our specific task and incorporate a word embedding function 
%\vspace{-\topsep}
\begin{equation}
%\vspace{-\topsep}
E({\rm\bf w}) = [E(w_1), E(w_2),...,E(w_N)]\in \mathbb{R}^{N\times d},
\end{equation} 
where the embedding $E(\cdot)$ comes from BERT, to map words into continuous representations, instead of operating discrete input directly. So the transformational step ${\rm\bf w} \rightarrow {\rm\bf x}_0$ can be described as
%\vspace{-\topsep}
\begin{equation}
%\vspace{-\topsep}
\label{eq-embedding}
q({\rm\bf x}_0|{\rm\bf w}) = E({\rm\bf w}),
\end{equation}
where ${\rm\bf x}_0\in \mathbb{R}^{\rm d}$ is the initial representation of continuous Diffusion. Inversely, the trainable function 
%\vspace{-\topsep}
\begin{equation}
%\vspace{-\topsep}
\label{eq-pw0}
p_\theta({\rm\bf w}|{\rm\bf x}_0) = \textstyle\prod_{i=1}^n p_\theta(w_i|x_i)
\end{equation}
is utilized to transfer continuous vectors into words. Among them, $x_i$ is the representation of $i$-th word in ${\rm\bf x}_0$ and $p_\theta(w_i|x_i)$ is an MLP network with softmax, mapping a high-dimensional $x_i$ to a specific token $w_i$. 

The model uses a Markov Chain $\{{\rm\bf x}_0,\ {\rm\bf x}_1,\ ..,\ {\rm\bf x}_t,\ ..,\ {\rm\bf x}_T \}$ to model the "noising" step and generate a Gaussian distribution ${\rm\bf x}_T \sim {\cal N}(0,{\rm\bf I})$. The forward noising process is parameterized by 
%\vspace{-\topsep}
\begin{equation}
%\vspace{-\topsep}
\label{eq-forward}
 q({\rm\bf x}_t|{\rm\bf x}_{t-1}) = {\cal N}({\rm\bf x}_t; \sqrt{1-\beta_t}{\rm\bf x}_{t-1}, \beta_t{\rm\bf I}) , 
\end{equation}
where $\beta_t$ is the amount of noise added in the $t$-th step of noising phase. ${\rm\bf x}_T$ is sampled as the initial state at the beginning of the reverse process, and the backward denoising can be formulated as 
%\vspace{-\topsep}
\begin{equation}
%\vspace{-\topsep}
\label{eq-backward}
p_\theta({\rm\bf x}_{t-1}|{\rm\bf x}_t) = {\cal N}({\rm\bf x}_{t-1}; \mu_\theta({\rm\bf x}_t,t),{\cal \sigma}_\theta({\rm\bf x}_t,t)), 
\end{equation}
where functions $\mu_\theta$ and $\sigma_\theta$ are learnable and trained in the reverse phase. %Thus, the training goal of the Diffusion model is to estimate the distribution of $p_\theta$ where the Variational Lower-Bound (VLB) is used as a computable lower bound. 

Based on the forward noising process (Eq.\ref{eq-forward}) and re-parameterizing trick, ${\rm\bf x}_t$ can be expressed by ${\rm\bf x}_0$:
%\vspace{-\topsep}
\begin{equation}
%\vspace{-\topsep}
\begin{aligned}
   {\rm\bf x}_t &= \sqrt{\alpha_t}{\rm\bf x}_{t-1}+\sqrt{1-\alpha_t}{\rm\bf z}_{t-1}\\
%       &= \sqrt{\alpha_t\alpha_{t-1}}{\rm\bf x}_{t-2}\\
%       &\qquad +\sqrt{1-\alpha_{t-1}}(\sqrt{\alpha_t}z_{t-2}+z_{t-1}) \\
       &= \sqrt{\tilde{\alpha}}{\rm\bf x}_0+\sqrt{1-\tilde{\alpha}}\tilde{\rm\bf z},
\end{aligned}
\end{equation}
where $\alpha_t = 1 - \beta_t$ and $\tilde{\alpha} = \prod_{t=1}^T \alpha_t$. In addition, noise added is defined by $ {\rm\bf z}_t\sim{\cal N}(0,{\rm\bf I}) $ and $\tilde{\rm\bf z}$ is the Gaussian superposition of $\{{\rm\bf z}_0,{\rm\bf z}_1, ..., {\rm\bf z}_t\}$.
%$ q({\rm\bf x}_t|{\rm\bf x}_0) = \prod_{t=1}^T q({\rm\bf x}_t|{\rm\bf x}_{t-1})$
%$ \sqrt{1-\alpha_{t-1}}(\sqrt{\alpha_t}z_{t-2}+z_{t-1}) = \sqrt{1-\alpha_t\alpha_{t-1}}\ \tilde{z}_{t-1}$

Therefore, the training goal of the Diffusion model, which is also regarded as the semantic loss function ${\cal{L}}_{\rm S}$, is to estimate the distribution of $p_\theta$ in which the VLB (Variational Lower-Bound) is used as a computable lower-bound:
%\vspace{-\topsep}
\begin{equation}
%\vspace{-\topsep}
\begin{aligned}
-\mathbb{E}_{q({\rm\bf x}_0)}[\log p_\theta({\rm\bf x}_0)]
&\leq \mathbb{E}_{q({\rm\bf x}_{0:T})}\bigg[\log \frac{q({\rm\bf x}_{1:T}|{\rm\bf x}_0)}{p_\theta({\rm\bf x}_{0:T})}\bigg] \\
&\textstyle= \mathbb{E}_{q({\rm\bf x}_{0:T})}\bigg[\frac{1}{2\sigma^2}|| \hat{\mu}({\rm\bf x}_T,{\rm\bf x}_0) ||^{2} \\
&\textstyle ~~ +\sum_{t=2}^T\frac{1}{2\sigma^2}|| \mu_\theta({\rm\bf x}_t,t) - \hat{\mu}({\rm\bf x}_t,{\rm\bf x}_0) ||^{2} \\
&\textstyle ~~ -\log p_\theta({\rm\bf x}_0|{\rm\bf x}_1)\bigg],
\end{aligned}
\end{equation}
where $\mu_\theta$ is the mean of $p_\theta({\rm\bf x}_{t-1}|{\rm\bf x}_t)$ and $\hat{\mu}$ is the mean of the posterior $q({\rm\bf x}_{t-1}|{\rm\bf x}_0, {\rm\bf x}_t)$. Removing the constant terms and the coefficient $\frac{1}{2\sigma^2}$, the loss function is simplified as:
%\vspace{-\topsep}
\begin{equation}
%\vspace{-\topsep}
\begin{aligned}
{\cal{L}}_{\rm S}({\rm\bf x}_0) = \textstyle\sum_{t=1}^T{\mathbb{E}}|| \mu_\theta({\rm\bf x}_t,t) - \hat{\mu}({\rm\bf x}_t,{\rm\bf x}_0) ||^{2}.
\end{aligned}
\end{equation}

Combined with the step ${\rm\bf w} \rightarrow {\rm\bf x}_0$ and ${\rm\bf x}_0 \rightarrow {\rm\bf w}$, the loss function can be rewritten as:
%\vspace{-\topsep}
\begin{equation}
%\vspace{-\topsep}
\begin{aligned}
{\cal L}_{\rm S}({\rm\bf w}) &= \mathbb{E}[{\cal L}_{\rm S}({\rm\bf x}_0)+\log q({\rm\bf x}_0|{\rm\bf w}) -\log p_\theta({\rm\bf w}|{\rm\bf x}_0)]\\
    % &= \mathbb{E}[|| \hat{\mu}({\rm\bf x}_T,{\rm\bf x}_0) ||^{2} +\sum_{t=2}^T|| \mu_\theta({\rm\bf x}_t,t) - \hat{\mu}({\rm\bf x}_t,{\rm\bf x}_0) ||^{2} \\
    % &\qquad +||E({\rm\bf w})-\mu_\theta({\rm\bf x}_1,1)||^{2} -\log p_\theta({\rm\bf w}|{\rm\bf x}_0)] \\
    &= \mathbb{E}[{\cal L}_{\rm S}({\rm\bf x}_0)+||E({\rm\bf w})-\mu_\theta({\rm\bf x}_1,1)||^{2} \\
    &\quad -\log p_\theta({\rm\bf w}|{\rm\bf x}_0)].
\end{aligned}
\end{equation}

\subsection{Metrical Controller}
\label{Metrical Controller}
\textbf{Motivation} \indent To ensure the primary model focuses more on the generation of text content itself, we devise a separately trained Metrical Controller to achieve format and rhyme control. In this way, PoetryDiffusion does not need to concatenate controlling encoding onto content-encoding as many previous methods do, which could scatter the model's semantic attention, thus minimizing the potential weakening of semantic representation caused by metrical control. Furthermore, the modular controller design enables our method to be easily adapted to other CTG tasks, significantly enhancing the practicality and versatility of our approach.

We employ deep neural network-based classifiers as the metrical controller due to two key advantages. Firstly, they adeptly model intricate distributions of specific attributes, thereby offering precise guidance during the diffusion process. Secondly, these classifiers enhance stability by easily calculating and imposing penalties on states that stray significantly from the target distribution. This not only addresses the instability often associated with the diffusion process but also ensures reliable samples.

\subsubsection{Format}

The chosen poetic forms, Sonnet and SongCi, exhibit considerable flexibility in terms of sentence length. For instance, SongCi's under different CiPai's feature distinct theatrical formats, which differ from the fixed 5-character and 7-character poetic structures. 
%The chosen poetic forms, Sonnet and SongCi, boast varying sentence lengths and specific theatrical formats dictated by CiPai.
Furthermore, while end signals of line or sentence are present in the original data, these signals encapsulate a significant amount of control information, encompassing not only line count and sentence length but also implicit positional cues for rhyme words, which must be at the end of lines or sentences. Therefore, explicit encoding of format information is essential to enhance format control and emphasize other associated details. %\chumin{Motivation for format control, \#3 Question B}

We define a sequence of format metrics, denoted by $S$ (with the same length as ${\rm\bf w}$), to indicate the target locations of ending signals. $S=(m,...,m,\langle eos \rangle,m,...)$, where "$\langle eos \rangle$" represents the end of each Sonnet line and $m$ is a mask symbol meaning that its corresponding word has no specific format rule. In SongCi, the punctuation characters "," and "." will replace "$\langle eos \rangle$" and act as the ending signals of each sentence. Similarly, $S$ can be generalized to other sentences with requirements of the signal's location.
The format loss is calculated using MSL (Mean-Squared Loss) between target sequence $S$ and predicted sequence ${\rm\bf x}_0$ based on the Diffusion feature representation ${\rm\bf x}_t$. The formula for the format loss is 
%\vspace{-\topsep}
\begin{equation}
%\vspace{-\topsep}
\label{eq-format}
{\cal L}_{\rm format}={\it MSL}(S,\varphi({\rm\bf x}_t)),
\end{equation}
where $\varphi(\cdot)$ is an MLP network with softmax.

\subsubsection{Rhyme}
Regarding rhyme control, we construct a rhyme categories space, whose representation is a vector in $\mathbb{R}^{\rm 6219}$ in Sonnet and $\mathbb{R}^{\rm 17}$ in SongCi. The last ($l$-th) word of each ($n$-th) line, $w_{n_l}$, is chosen as the input for the word-level classifier based on BERT \cite{devlin2018bert}. This classifier aims to provide the convincing rhyme category of $w_{n_l}$ as its output. Additionally, the tone rule constraints of all words, $w_{m}$, in SongCi should also be considered, with a tone categories space of $\mathbb{R}^{\rm 3}$ ("+", "-", "0"). 
After acquiring the representation of $w_{n_l}$ or $w_{m}$, we can readily compute the probability distribution of rhyme or tone rules by applying an MLP network with softmax. Hence, the loss can be formulated as 
%\vspace{-\topsep}
\begin{align}
%\vspace{-\topsep}
\label{eq-rhyme}
{\cal L}_{\rm rhyme} &= \rho_{n_l}\log({\it BERT}(w_{n_l}; \rho_{n_l})),\\
\label{eq-tone}
{\cal L}_{\rm tone} &= \tau_{m} \log({\it BERT}(w_{m}; \tau_{m})), 
\end{align}
where $\rho_{n_l}$ and $\tau_{m}$ are rhyme and tone ground truth labels of the target word. 

Notably, the Controller is employed throughout the denoising process, rather than solely in the final step, thereby achieving concurrent augmentation of semantics and metrics. Consequently, the format and rhyme of the poetry being generated will progressively enhance amid semantic refinement, avoiding any detriment to the meticulously crafted semantics in the end.%\chumin{\#2 Other suggestions 3}

\subsection{Joint Manipulation}
\label{joint manipulation}
The controllable decoding process is similar to the process of training the Diffusion model. While, the distinction lies in the use of a trained $p_\theta$ as the initial denoising distribution and we conduct sampling from ${\rm \bf x}_T$ to ${\rm \bf x}_0$.

When we combine PoetryDiffusion and Metrical Controller to generate poetry, the feature representation from each step of Diffusion would be adopted to act as the input of Metrical Controller. In step $i-1$, sample ${\rm \bf x}_{i-1}$ through $p_\theta$ (Eq.\ref{eq-backward}), input ${\rm \bf x}_{i-1}$ and condition $\rho$ into the well-trained BERT-based Metrical Controller to obtain the metric loss (Eq.\ref{eq-format}-\ref{eq-tone}). Then, the Controller transmits the Metrical loss ${\cal L}_{\rm M}$ to PoetryDiffusion, which can be written as
%\vspace{-\topsep}
\begin{equation}
%\vspace{-\topsep}
    {\cal L}_{\rm M} = \lambda_1{\cal L}_{\rm format} +\lambda_2{\cal L}_{\rm tone} + \lambda_3{\cal L}_{\rm rhyme},
\end{equation}
where $\lambda_i\ (i=1,2,3)$ are the hyperparameters selected by the scale of semantic loss and losses for each metric, ensuring that metrical losses can impact the performance evenly without overshadowing semantic aspects. The term related to ${\cal L}_{\rm tone}$ should be omitted when dealing with sonnets.

The final loss of each step which would affect the denoising process, updating $p_\theta$, is:
%\vspace{-\topsep}
\begin{equation}
%\vspace{-\topsep}
{\cal L} = {\cal L}_{\rm S} + {\cal L}_{\rm M}.
\end{equation}
Then ${\rm \bf x}_{i}$ would be sampled based on new $p_\theta$. 

Consequently, the feature representation would be determined by PoetryDiffusion and its Metrical Controller through the loss ${\cal L}$ in each step. This process continues until we sample ${\rm \bf x}_0$ and decode it to ${\rm \bf w}$ through Eq.\ref{eq-pw0}.

\begin{table*}[ht]
% \scriptsize
% \renewcommand{\arraystretch}{0.75}
\small
% %\vspace{-\topsep}
\centering
\begin{tabular}{l|ccccc|ccc|c} 
\toprule
    \multirow{2}{*}{\bf Model} & \multicolumn{5}{c}{\bf Semantics} & \multicolumn{3}{|c|}{\bf Metrics} & \multirow{2}{*}{\bf Overall $\uparrow$} \\ 
    & B{\tiny LEU} $\uparrow$ & R{\tiny OUGE} $\uparrow$ & Distinct $\uparrow$ & PPL $\downarrow$ & Avg $\uparrow$ & Format $\uparrow$ & Rhyme $\uparrow$ & Avg $\uparrow$ &\\
\midrule
    SeqGAN      & 26.56 & 27.61 & 82.24 & 32.93 & 50.87 & 97.13 & 35.41 & 66.27 & 56.00\\
    MRCG		& 28.18  & 23.63 & 55.14 & 13.04 & 48.48 & \textbf{100.00} & 37.59 & 68.80 & 55.25 \\ 
    SongNet    & 25.09 & 37.78 & 77.20 & 12.50 & 56.89 & 99.95 & 29.79 & 64.87 & 59.55 \\ 
    GPT3		& 26.59 & 32.70 & 59.01 & 11.72 & 51.65 & 75.63 & 35.55  & 55.59 & 52.96 \\ 
    ChatGPT		& 30.91 & 42.78 & 81.64 & 9.52 & 61.45 & 89.55 & 50.45 & 70.00  & 64.30 \\ 
    Llama2-70B-chat &32.20 & 41.45 & \textbf{87.30} & 8.72 & 63.06 & 96.96 & \textbf{54.64} & 75.80 & 67.31  \\ 
    PoetryDiffusion(w/o C)	    & 30.18  & 38.67 & 86.43 & \textbf{8.48} & 61.70 & 96.00 & 23.68  & 59.84 & 61.08 \\ 
    PoetryDiffusion	 & \textbf{32.94} & \textbf{44.75} & 87.15 & 10.44 & \textbf{63.60} & \textbf{100.00} & 52.28 & \textbf{76.14} & \textbf{67.78}\\
\bottomrule
\end{tabular}
%\vspace{-\topsep}
\caption{Performance on {\bf Sonnet} obtained by the testing methods. The best results are in \textbf{bold}.}
%\vspace{-\topsep}
\label{Sonnet Performance}
\end{table*}

\begin{table*}[ht] 
% \scriptsize
% \renewcommand{\arraystretch}{0.75}
\small
%\vspace{-\topsep}
\centering
\begin{tabular}{l|ccccc|cccc|c} 
\toprule
    \multirow{2}{*}{\bf Model} & \multicolumn{5}{c}{\bf Semantics} & \multicolumn{4}{|c|}{\bf Metrics} & \multirow{2}{*}{\bf Overall $\uparrow$} \\
    & B{\tiny LEU} $\uparrow$ & R{\tiny OUGE} $\uparrow$ & Distinct $\uparrow$ & PPL $\downarrow$ & Avg $\uparrow$ & Format $\uparrow$ & Tone $\uparrow$ & Rhyme $\uparrow$ & Avg $\uparrow$ & \\
\midrule
    SeqGAN      & 24.49 & 15.45 & 90.06 & 10.79 & 55.30 & 79.58 & 65.68 & 53.77 & 66.27 & 60.78\\
    MRCG		& 22.90  & 14.78  & 90.06 & 10.32 & 54.42 & 99.35 & \textbf{93.71} & \textbf{98.28} & \textbf{97.02} & 75.72\\ 
    SongNet   & 21.23 & 14.04  & 86.82 & 11.48 & 52.65 & 99.42 & 76.22 & 80.01 & 85.22 & 68.93\\ 
    GPT3   & 25.17 & 16.17 & 71.88 & 9.77 & 50.86 & 71.80 & 50.13 & 29.64 & 50.52 & 50.69\\ 
    ChatGPT   & 18.29 & 11.96 & 91.36 & 8.79 & 53.20 & 84.58 & 70.23 & 51.55 & 68.79 & 61.00\\ 
    Llama2-70B-chat &23.79 & 15.41 & 90.70 & 10.28 & 54.91 & 80.03 & 65.11 & 51.98 & 65.71 & 59.53 \\
    PoetryDiffusion(w/o C)	 & 25.59 & 16.86 & 92.06 & 9.14 & 56.35 & 80.44 & 64.33 & 50.94 & 65.24 & 60.79\\ 
    PoetryDiffusion		 & \textbf{28.98} & \textbf{17.11}  & \textbf{92.07} & \textbf{8.76} & \textbf{57.35} & \textbf{99.51} & 91.64 & 95.37 & 95.51 & \textbf{76.43} \\ 
\bottomrule
\end{tabular}
%\vspace{-\topsep}
\caption{Performance on {\bf SongCi} obtained by the testing methods. The best results are in \textbf{bold}.} 
%\vspace{-3mm}
\label{SongCi Performance}
\end{table*}

\begin{table}[t] 
% %\vspace{-\topsep}
\small
\centering
\begin{tabular}{cc} 
\toprule
    \textbf{Type} & \textbf{Rhyme Scheme} \\
\midrule
    Shakespearean Sonnets & ABAB CDCD EFEF GG \\
    Spenserian Sonnets & ABAB BCBC CDCD EE \\
    Italian or Petrarchan Sonnets (1) & ABBA ABBA CDC CDC \\ 
    Italian or Petrarchan Sonnets (2) & ABBA ABBA CDE CDE \\
    Terza Rima Sonnet & ABA BCB CDC DED EE \\
\bottomrule
\end{tabular}
%\vspace{-\topsep}
\caption{Five types of sonnets and relevant rhyme schemes}
%\vspace{-5mm}
\label{sonnet types}
\end{table}

\section{Experiments}

\subsection{Dataset and Evaluation}
We train our model on two datasets, Sonnet and SongCi. Sonnet consists of 3,355 sonnets collected by \cite{lau2018deep}. SongCi comprises 82,724 SongCi’s, curated by \cite{zhang2019generating}. 

To evaluate semantic and metrical performance together, we propose a simple average evaluation score: 
%\vspace{-\topsep}
\begin{equation*}
%\vspace{-\topsep}
\begin{aligned}
S^{\rm Sonnet}_{\rm overall} &= 0.5\times {\it avg}(S_{\rm BLEU},S_{\rm ROUGE},S_{\rm Distinct}, \\
    &\quad 100-S_{\rm PPL})+0.5\times {\it avg}(S_{\rm format},S_{\rm rhyme}),
\end{aligned}
\end{equation*}
where the previously settings for BLEU~\cite{papineni2002bleu}, ROUGE~\cite{lin2004rouge}, Distinct~\cite{li2015diversity} and Perplexity (PPL) are utilized. BLEU and ROUGE are scored by comparing generated poems, which are segmented into lines or subphases, to a reference database of sub-sentences with poetic phrases. PPL is computed by output, using the language-specific BERT. In addition, the tone accuracy in SongCi would be considered:
%\vspace{-\topsep}
\begin{equation*}
%\vspace{-\topsep}
\begin{aligned} 
S^{\rm SongCi}_{\rm overall} &= 0.5\times {\it avg}(S_{\rm BLEU},S_{\rm ROUGE},S_{\rm Distinct}, \\
    &100-S_{\rm PPL})+0.5\times {\it avg}(S_{\rm format},S_{\rm tone},S_{\rm rhyme}). 
\end{aligned}
\end{equation*}
% In the context of Sonnet, the format score $S_{\rm format}$ denotes the precision of line count accuracy. Conversely, for SongCi, it encapsulates the precision of both line count and individual line length. As for the rhyme score $S_{\rm rhyme}$ in Sonnet, it draws from a selection of classic sonnet forms (refer to Table~\ref{sonnet types}) to determine the best match. In the case of SongCi, the rhyme score takes into consideration words placed at specific positions that rhyme within the target poem. Elaborated methodologies pertaining to the computation of metrical scores can be found in Appendix~\ref{Appendix Eval Detail}A.

Moreover, more detailed methods of calculating metrical scores are described as follows.

\noindent \textbf{Format.} For Sonnet, the accuracy score (\%) is formulated as:
%\vspace{-\topsep}
\begin{equation*}
%\vspace{-\topsep}
S_{\rm format}=1-|{\it N}-14|/14,
\end{equation*}
where {\it N} stands for the number of lines in generated poetry and 14 is the fixed number of lines for sonnets. For SongCi, the formula is expressed as:
%\vspace{-\topsep}
\begin{equation*}
%\vspace{-\topsep}
S_{\rm format}=\it Ts/\it L,
\end{equation*}
where {\it Ts} stands for the number of symbols with the true type (ending marks or meaningful words) compared with the original poetry, and {\it L} is the whole length of the poetry.

\noindent \textbf{Rhyme.} For Sonnet, we try to match the rhyme scheme of each generated poetry with 5 types of classic sonnets (Table~\ref{sonnet types}) and report the highest accuracy score. The selected words for evaluation are the last words of each line. For SongCi, since not all the last words of sentences which end with "," or "." satisfy the same type of rhyme, we select the rhyme appearing most in the target original poetry and record their locations for evaluation. Words with the same rhyme on selected locations are regarded as true. The accuracy score (\%) of rhyme can be written as:
%\vspace{-\topsep}
\begin{equation*}
%\vspace{-\topsep}
S_{\rm rhyme}=\it Tr/\it Ls,
\end{equation*}
where {\it Tr} means the number of words with true rhyme within locations selected, and $Ls$ means the number of locations selected. Likewise, the accuracy score (\%) of tone can be expressed as:
%\vspace{-\topsep}
\begin{equation*}
%\vspace{-\topsep}
S_{\rm tone}=\it Tt/\it L,
\end{equation*}
where {\it Tt} means the number of words with the true tone, and {\it L} remains consistent with the previous statement.

\begin{table*}[htb] 
\small
\renewcommand{\arraystretch}{0.9}
%\vspace{-\topsep}
\centering
\begin{tabular}{l|cc|cc|cc|cc|cc} 
\toprule
    \multirow{2}{*}{\bf Model} & \multicolumn{2}{c}{\bf Fluency} & \multicolumn{2}{|c|}{\bf Coherence} & \multicolumn{2}{c}{\bf Meaningfulness} & \multicolumn{2}{|c}{\bf Poeticness} & \multicolumn{2}{|c}{\bf Average}\\
    & Sonnet & SongCi & Sonnet & SongCi & Sonnet & SongCi & Sonnet & SongCi & Sonnet & SongCi \\
% \midrule
%     Human Poet   & 3.60 & 3.69 & 3.40 & 3.59 & 3.24 & 3.78 & 3.88 & 3.74 & 3.53 & 3.70\\ 
\midrule
    SongNet   & 2.67 & 3.39 & 2.72 & 3.29 & 2.55 & 3.40 & 3.04 & 3.49 & 2.75 & 3.39\\ 
    ChatGPT & \textbf{3.45} & \textbf{3.46} & 3.35 & 3.29 & \textbf{3.60} & \textbf{3.52} & 3.19 & 3.20 & 3.40 & 3.37\\ 
    PoetryDiffusion & 3.40 & 3.43 & \textbf{3.43} & \textbf{3.44} & 3.32 & 3.47 & \textbf{3.62} & \textbf{3.52} & \textbf{3.43} & \textbf{3.46}\\ 
\bottomrule
\end{tabular}
%\vspace{-\topsep}
\caption{Mean ratings elicited by humans on generated poetry. Best rates except for Human Poet are in \textbf{bold}.} 
%\vspace{-3mm}
\label{Human Eval}
\end{table*}

\subsection{Training Details}
This section shows the optimal hyperparameters of our PoetryDiffusion model. The number of decoding or encoding steps $T$ is set to be 2000 steps. In addition, we rescale the diffusion steps into 200 to accumulate the poetry generation process based on DDIM \cite{song2020denoising}. The dimension of word embedding is chosen to be 16. The method of organizing batches differs between the two datasets. For Sonnet, pad each piece of poetry to the same length and then concatenate the number of sequences corresponding to batch size. While for SongCi, firstly concatenate all sequences of text and then cut into blocks with appropriate shapes. The number of training iterations is set to 150K. It takes approximately 4 hours to train PoetryDiffusion and Metrical Controller on an NVIDIA A100 GPU monopolized by one job.

\subsection{Compared Prior Art}
We conduct a comparative analysis between our proposed method and established state-of-the-art (SOTA) techniques. To ensure a fair comparison, datasets in two languages are partitioned into train/valid/test in the same way as used in previous work. Details of the realization are listed below.
% \textbf{SeqGAN}  employs a GAN framework, treating the generator as a stochastic policy in reinforcement learning. %It back-propagates rewards from the discriminator to encourage the generator in producing poetic text resembling human composition. 
% \textbf{MRCG} \cite{zhang2019generating} introduces a CVAE framework to generate SongCi while adhering to metric constraints. %Notably, it is the pioneering model to explicitly encode metrical restrictions within a neural architecture. 
% \textbf{SongNet} \cite{li2020rigid} integrates metrical symbols into continuous representations and combines them with a Transformer-based autoregressive language model.  
% Additionally, we present results from experiments involving fine-tuned \textbf{GPT3} and one-shot \textbf{ChatGPT} for comprehensive evaluation.

\paragraph{SeqGAN} \cite{yu2017seqgan} employs a GAN framework, treating the generator as a stochastic policy in reinforcement learning. %It back-propagates rewards from the discriminator to encourage the generator in producing poetic text resembling human composition.
We utilize the inherent approach in its unaltered form to accomplish the task while substituting our dataset.

\paragraph{MRCG} \cite{zhang2019generating} introduces a CVAE framework to generate SongCi while adhering to metric constraints. %Notably, it is the pioneering model to explicitly encode metrical restrictions within a neural architecture. 
When generating SongCi, we simply follow its method and settings. And we migrate the model to the Sonnet dataset by changing the Chinese rhyme rules into English and removing the restriction of tone.

\paragraph{SongNet} \cite{li2020rigid} integrates metrical symbols into continuous representations and combines them with a Transformer-based autoregressive language model. We directly employ the original method to complete the task, with our dataset replaced.

\paragraph{GPT3} \cite{brown2020language} is fine-tuned on SongCi and Sonnet respectively. For the SongCi dataset, CiPai, which can be regarded as the title of SongCi, acts as the prompt to generate a whole poem. However, due to the lack of titles in the Sonnet dataset, GPT3 receives the first line content in Sonnet as its prompt to generate the rest of the poetry.

\begin{figure}[htb]
%\vspace{-3mm}
    \centering
    \includegraphics[width=1\linewidth]{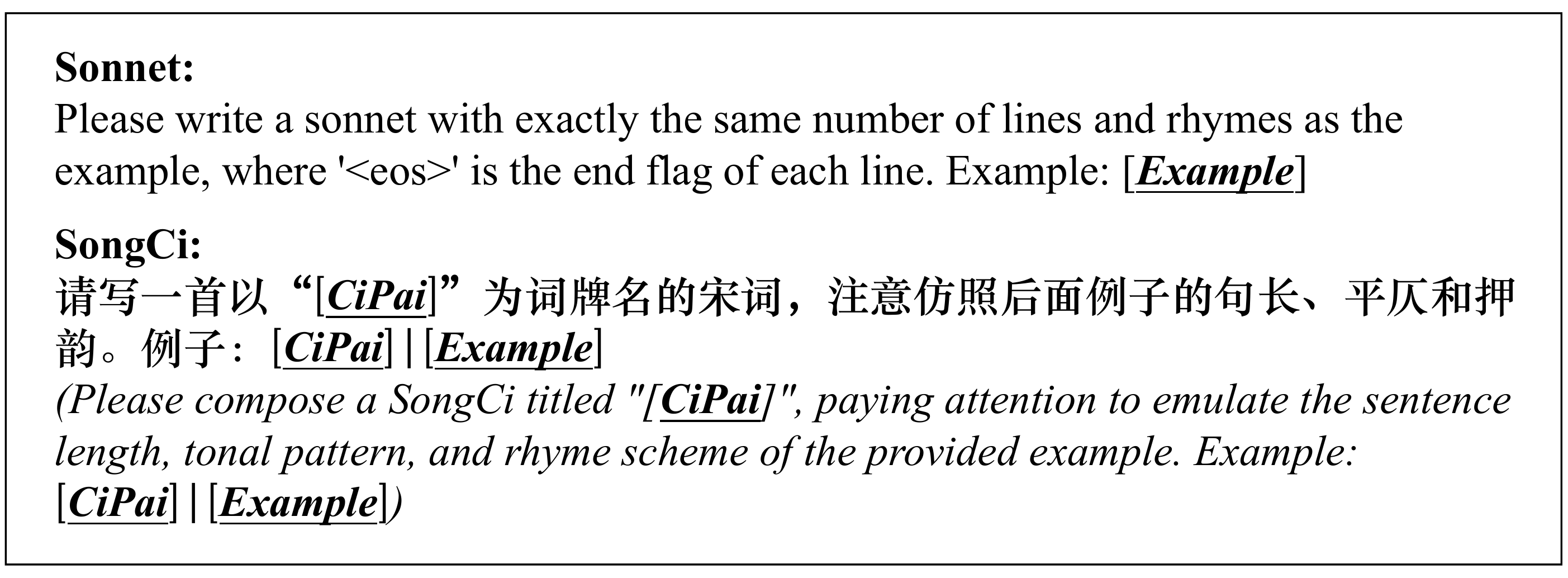}
    %\vspace{-5mm}
    \caption{Prompts for ChatGPT to generate poetry}
    %\vspace{-5mm}
    \label{prompt}
\end{figure}

\paragraph{ChatGPT} \cite{openai2021gpt35turbo} (GPT-3.5-Turbo) is asked to generate a new Sonnet or SongCi under one instruction example in the test set. Prompts used are as Figure \ref{prompt}.

\paragraph{Llama2-70B-chat} \cite{touvron2023llama} is the open source LLM developed by meta. We utilize all the SongCi and Sonnet training data to finetune Llama2 based on the LoRA technique \cite{hu2021lora}. During finetuning, we employ similar format in Figure~\ref{prompt}, which use the poetry generation instruction and examples as the input, and then enable Llama2 to generate the poetry conforming to the corresponding format.

%\zhiyuan{the specific version of gpt3 and chatgpt should be added}

\begin{figure*}[htb]
%\vspace{-\topsep}
    \centering
    \includegraphics[width=1\textwidth]{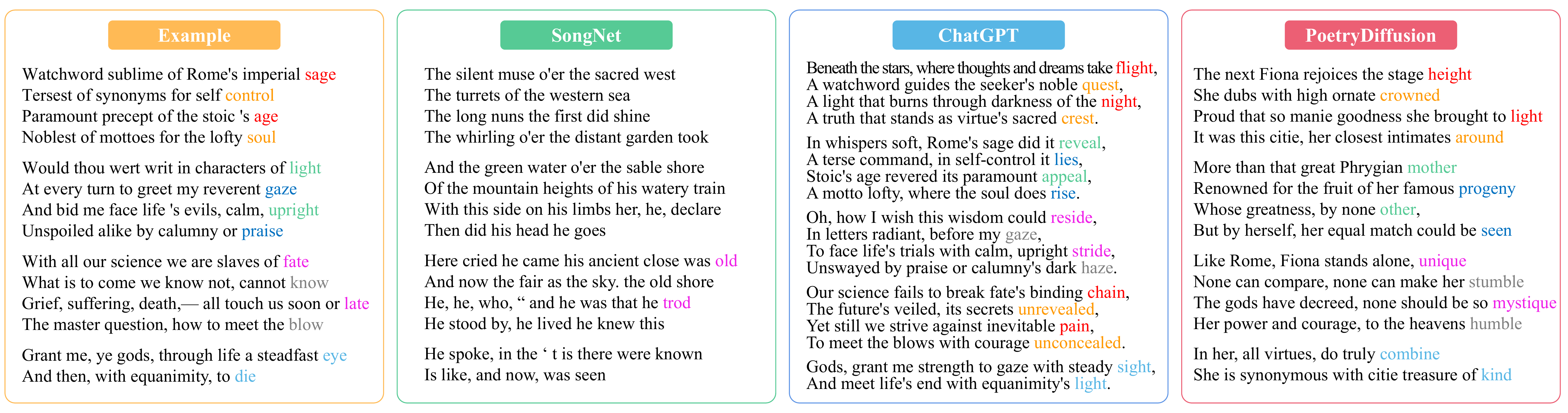}
    \vspace{-5mm}
    \caption{Sonnets generated by models given the same example. Pairs of words with the same color demonstrate accurate rhyme, as they share a common rhyme scheme.}
    \vspace{-5mm}
    \label{Sonnet Case}
\end{figure*}

\begin{figure}[htb]
% %\vspace{-\topsep}
    \centering
    \includegraphics[width=1\linewidth]{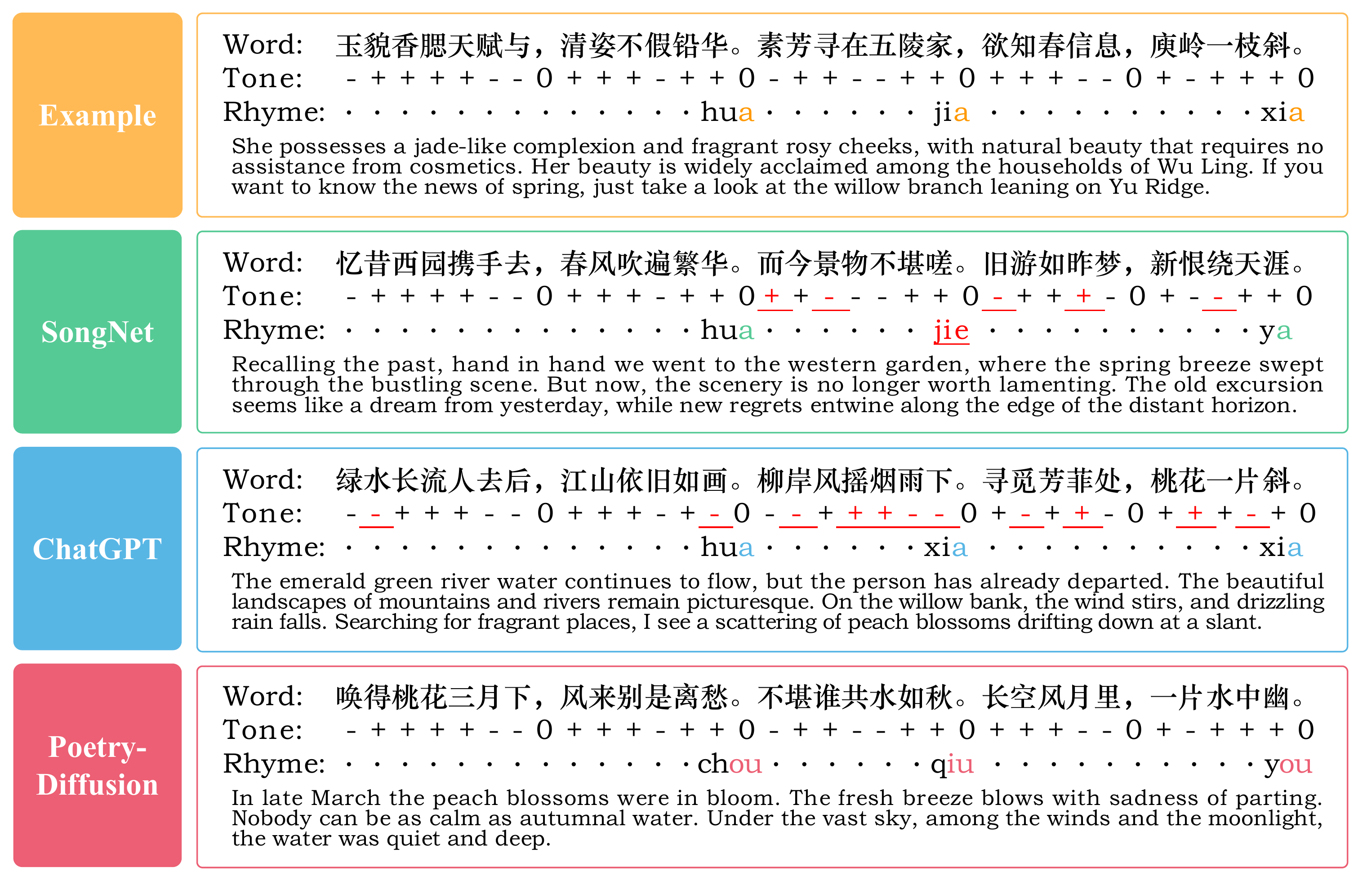}
    %\vspace{-5.5mm}
    \caption{SongCi's generated by models given the same example. Errors in Tone and Rhyme control are indicated using both red font and underlining.}
    \vspace{-3mm}
    \label{Songci Case}
\end{figure}

\subsection{Experimental Performance}
\subsubsection{Automatic Evaluation}
With a focus on semantic performance, as shown in Table~\ref{Sonnet Performance} and Table~\ref{SongCi Performance}, PoetryDiffusion outperforms other models on both types of poetry, offering strong evidence of the efficacy of our model on semantic enhancement. It demonstrates the superior performance than the auto-regressive model like SongNet and LLMs such as ChatGPT and finetuned Llama2. 
As for metrical performance, PoetryDiffusion achieves the new SOTA results about average performance in Sonnet, surpassing the baselines with an obvious margin. It must be noted however that PoetryDiffusion's metrics are slightly worse than MRCG in SongCi. Further dataset analysis reveals SongCi demands more rigorous and intricate metrics. The rigid yet impactful final forced-word replacement technique in MRCG contributes to its favorable metrics but compromised semantics.
In conclusion, the SOTA overall score proves that our model simultaneously performs well on both semantic and metrical sides compared with all kind of baseline models.
%\textcolor{red}{add more detailed experimental explanation}

\subsubsection{Ablation Study}
%Taking the ablation study into account, t
The semantic performance of PoetryDiffusion(w/o \textbf{C}ontroller) is among the best across baseline models, demonstrating the superiority of the Diffusion model in text generation, supporting our choice of it. 
%Remarkably, due to diverse metrical information guiding the model to generate consecutive poetic phrases closely aligned with the writing characteristics of poetry, PoetryDiffusion (w/o \textbf{C}) slightly underperforms the full version. 
Remarkably, when combining the metrical controller, we can obtain further improvement in semantic performance. Metrics capture the essence of a poem's rhythm and sound, and by incorporating metrical controllers, we maintain the authentic emulation of prosody, enhancing semantic expression through structured poetic patterns that shape the composition of the text.
Meanwhile, compared to the full PoetryDiffuion, the metrical performance of the one without the Controller has a significant decrease, providing evidence that the controller is vital for augmenting metrical abilities.

\subsection{Human Evaluation}
\subsubsection{Criteria}
We employ the assessment methodology introduced by \citet{zhang2014chinese}, where human annotators rate poems using a 1-5 scale across four key dimensions:
\begin{itemize}
    \item \textit{Fluency}: be grammatical and syntactically well-formed
    \item \textit{Coherence}: be thematically structured
    \item \textit{Meaningfulness}: convey a meaningful message
    \item \textit{Poeticness}: display the features of a poem
\end{itemize}
% {\it Fluency}, {\it Coherence}, {\it Meaningfulness} and {\it Poeticness}. 

\subsubsection{Baselines}
To avoid aesthetic fatigue from manually reviewing extensive poetry, we selected a few models for human evaluation. We chose SongNet, with near-top scores in both datasets, to represent autoregressive models because it's tested on both Chinese and English poetry in the original paper \cite{li2020rigid}. ChatGPT, known for its generative prowess, was picked to compare our approach with a leading Large Language Model. 

\subsubsection{Settings}
All models are provided with the same examples to produce 25 Sonnets and 25 SongCi's. A panel of 5 experts, proficient university students who have majored in English and Chinese literature respectively, assesses the generated poems, and the average of their rating scores is used as the ultimate evaluation score.

\subsubsection{Result}
As shown in Table~\ref{Human Eval}, our PoetryDiffusion surpasses all baseline models in overall average scores. It closely rivals the performance of ChatGPT, with ChatGPT even outperforming PoetryDiffusion in {\it Fluency} and {\it Meaningfulness}. This discrepancy can be attributed to ChatGPT having access to a significantly larger training dataset compared to ours, rendering it more adept at generating general conversational text, which places a strong emphasis on fluency and meaningfulness. Conversely, in other dimensions, {\it Coherence} and {\it Poeticness}, PoetryDiffusion excels over other models, producing text that exhibits a more distinct poetic style, aligning well with the objectives of our poetry generation task.%\chumin{Analysis of human eval}

\subsection{Case Study}
In Figure \ref{Sonnet Case} and \ref{Songci Case}, we compare the poetry generated by SongNet, ChatGPT, and our PoetryDiffusion to better illustrate our motivation.

In terms of Sonnet generation, SongNet could not achieve rhyme and ChatGPT exhibited deficiencies in line count control; For SongCi, both SongNet and ChatGPT exhibited slight inaccuracies in Tone and Rhyme control. In comparison, PoetryDiffusion successfully generated Sonnet and SongCi with precise control over format and rhyme\footnote{The vowels "ou" and "iu" in the Chinese phonetic alphabet rhyme with each other.}. Moreover, PoetryDiffusion demonstrates superior semantic attributes. Most intuitively, it exhibits enhanced diversity, in stark contrast to the consistent repetition of initial words observed in the sonnets generated by the other two models.

\section{Visualization of Stepwise Optimization}

\begin{figure}[htb]
%\vspace{-3mm}
    \centering
    \includegraphics[width=0.9\linewidth]{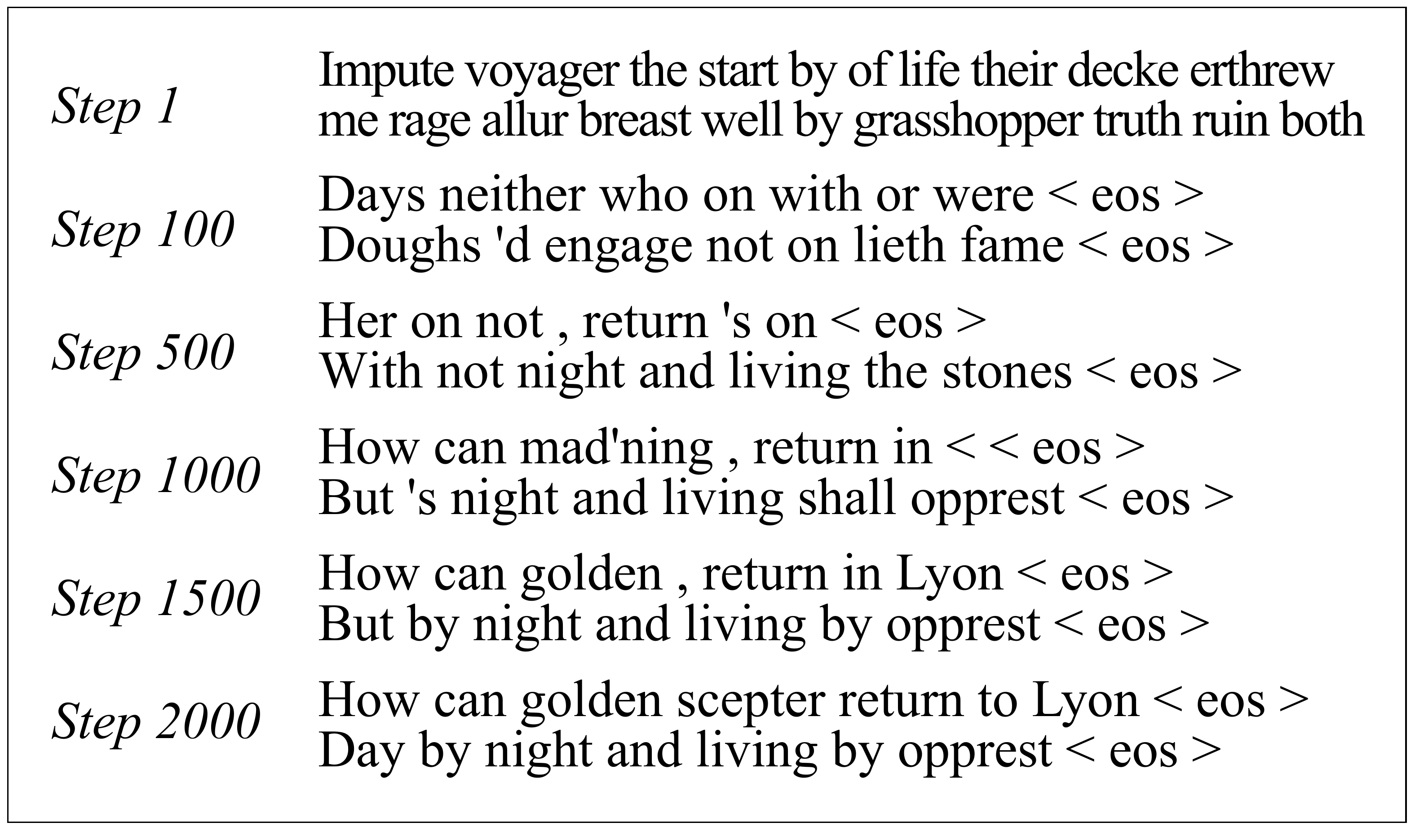}
    %\vspace{-\topsep}
    \caption{Generated Sonnet in Different Steps}
    %\vspace{-2mm}
    \label{Analysis Case}
\end{figure}
\begin{figure}[htb]
%\vspace{-\topsep}
    \centering
    \includegraphics[width=0.47\textwidth]{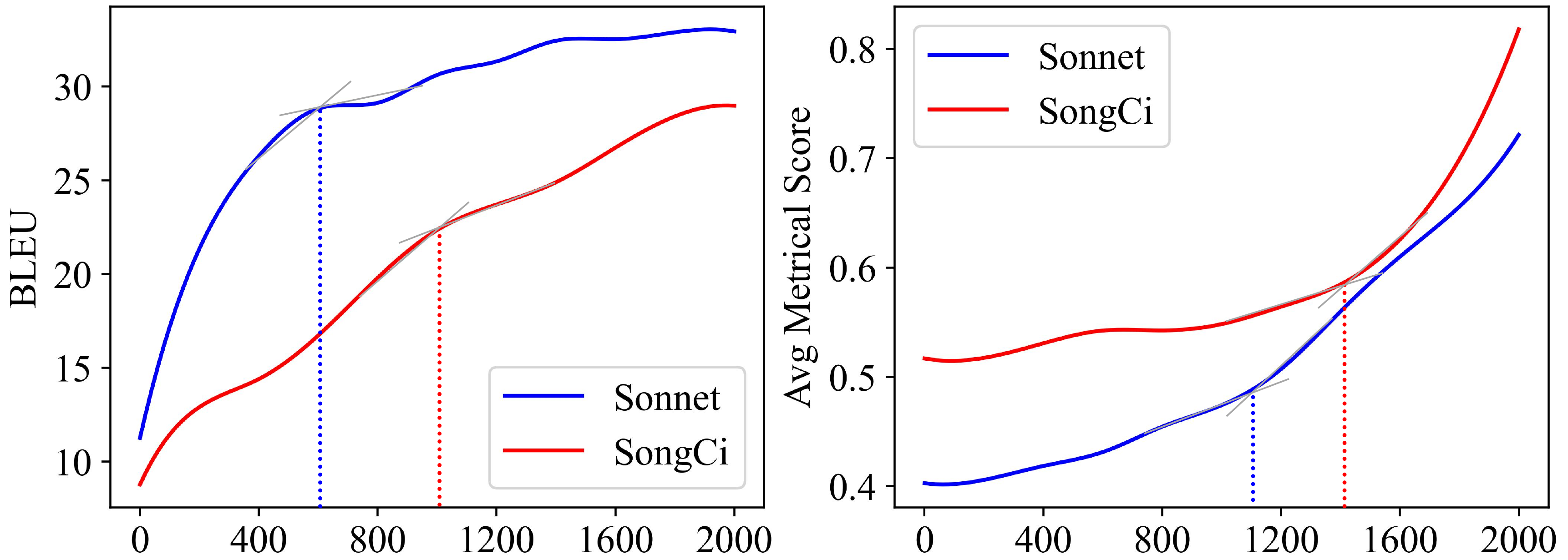}
    \vspace{-3mm}
    \caption{Stepwise evaluation scores. The intersection of the gray dashed lines highlights the point where the curve's rate of improvement changes.}
    \vspace{-4mm}
    \label{curve1}
\end{figure}
The denoising process may serve as a mechanism to stepwise predict the masked semantic and metrical information. To evaluate its assumption and reveal how PoetryDiffusion integrates semantics and metrics gradually, we conducted experiments focusing on poetry expression, BLEU, and metrical score stepwise.

% \begin{figure}[htb]
% %\vspace{-\topsep}
%     \centering
%     \includegraphics[width=0.47\textwidth]{Pic/Curve2_1.pdf}
%     \caption{Scores Curve in Different Steps}
%     \label{curve2}
% \end{figure}

As seen in Figure~\ref{Analysis Case}, the denoising process leads to a clearer topic, improved fluency, and a reduction in hallucinations in the later steps.
Furthermore, as depicted by the plotted curve (Figure~\ref{curve1}), the BLEU score exhibits a consistent upward trend, reaching its peak at the end of the steps. It is noteworthy that this upward trend is discernible at the onset and subsequently moderating in the first half of the steps. In contrast, the metrical score displays an initial gradual ascent, followed by an obvious acceleration in the latter half of the steps. 

These findings suggest that the proposed model establishes thematic semantics first, and it is only as the theme becomes relatively distinct that the influence of metrical control becomes more pronounced. This process steers the model towards imbuing metrical control into the poetry while upholding its semantic structure. These findings also elucidate the reason our PoetryDiffusion outperforms other generative models in terms of both semantics and metrics.

\section{Discussion}

In terms of the generalization of our method in other controllable text generation scenarios, we selected poetry generation for evaluation because its strict structures make it perfect for testing controlled text generation. The main challenges are ensuring the model to meet specific conditions and quantitatively evaluating the method's controllability. Poetry, with its well-defined rules, meets these challenges better than topic control, which is often too vague. If the model excels in poetry's stringent conditions, it likely will perform well in more relaxed contexts.

\section{Conclusion}
We proposed PoetryDiffusion which optimizes the semantic performance by stepwise denoising masked information in entire sentences and incorporating an exceptional metrical controller. By jointly utilizing these two components to generate poetry, a harmonious blend of semantic expression and syntactic control is achieved. SOTA performances in the automatic evaluation and human evaluation of PoetryDiffusion in two datasets also validate its effectiveness. Moreover, the cases study vividly showcases our model's superiority, and the visualization of the stepwise optimization process in the Diffusion model uncovers the different modeling phases of semantic features and metrical information.

\newpage

\bibliography{aaai24.bib}

\begin{thebibliography}{37}
\providecommand{\natexlab}[1]{#1}

\bibitem[{Benhardt et~al.(2018)Benhardt, Hase, Zhu, and
  Rudin}]{benhardt2018shall}
Benhardt, J.; Hase, P.; Zhu, L.; and Rudin, C. 2018.
\newblock Shall i compare thee to a machine-written sonnet? an approach to
  algorithmic sonnet generation.
\newblock \emph{arXiv preprint arXiv:1811.05067}.

\bibitem[{Betti, Ramponi, and Piccardi(2020)}]{betti-etal-2020-controlled}
Betti, F.; Ramponi, G.; and Piccardi, M. 2020.
\newblock Controlled Text Generation with Adversarial Learning.
\newblock In \emph{Proceedings of the 13th International Conference on Natural
  Language Generation}, 29--34. Dublin, Ireland: Association for Computational
  Linguistics.

\bibitem[{Bond-Taylor et~al.(2022)Bond-Taylor, Leach, Long, and
  Willcocks}]{9555209}
Bond-Taylor, S.; Leach, A.; Long, Y.; and Willcocks, C.~G. 2022.
\newblock Deep Generative Modelling: A Comparative Review of VAEs, GANs,
  Normalizing Flows, Energy-Based and Autoregressive Models.
\newblock \emph{IEEE Transactions on Pattern Analysis and Machine
  Intelligence}, 44(11): 7327--7347.

\bibitem[{Brown et~al.(2020)Brown, Mann, Ryder, Subbiah, Kaplan, Dhariwal,
  Neelakantan, Shyam, Sastry, Askell et~al.}]{brown2020language}
Brown, T.; Mann, B.; Ryder, N.; Subbiah, M.; Kaplan, J.~D.; Dhariwal, P.;
  Neelakantan, A.; Shyam, P.; Sastry, G.; Askell, A.; et~al. 2020.
\newblock Language models are few-shot learners.
\newblock \emph{Advances in neural information processing systems}, 33:
  1877--1901.

\bibitem[{Che et~al.(2017)Che, Li, Zhang, Hjelm, Li, Song, and
  Bengio}]{che1702maximum}
Che, T.; Li, Y.; Zhang, R.; Hjelm, R.~D.; Li, W.; Song, Y.; and Bengio, Y.
  2017.
\newblock Maximum-likelihood augmented discrete generative adversarial networks
  (2017).
\newblock \emph{arXiv preprint arXiv:1702.07983}.

\bibitem[{Chen et~al.(2019)Chen, Yi, Sun, Li, Yang, and
  Guo}]{chen2019sentiment}
Chen, H.; Yi, X.; Sun, M.; Li, W.; Yang, C.; and Guo, Z. 2019.
\newblock Sentiment-Controllable Chinese Poetry Generation.
\newblock In \emph{IJCAI}, 4925--4931.

\bibitem[{Chen and Yang(2023)}]{chen-yang-2023-controllable}
Chen, J.; and Yang, D. 2023.
\newblock Controllable Conversation Generation with Conversation Structures via
  Diffusion Models.
\newblock In \emph{Findings of the Association for Computational Linguistics:
  ACL 2023}, 7238--7251. Toronto, Canada: Association for Computational
  Linguistics.

\bibitem[{Deng et~al.(2020)Deng, Wang, Liang, Chen, Xie, Zhuang, Wang, and
  Xiao}]{deng2020iterative}
Deng, L.; Wang, J.; Liang, H.; Chen, H.; Xie, Z.; Zhuang, B.; Wang, S.; and
  Xiao, J. 2020.
\newblock An iterative polishing framework based on quality aware masked
  language model for Chinese poetry generation.
\newblock In \emph{Proceedings of the AAAI conference on artificial
  intelligence}, volume~34, 7643--7650.

\bibitem[{Devlin et~al.(2018)Devlin, Chang, Lee, and
  Toutanova}]{devlin2018bert}
Devlin, J.; Chang, M.-W.; Lee, K.; and Toutanova, K. 2018.
\newblock Bert: Pre-training of deep bidirectional transformers for language
  understanding.
\newblock \emph{arXiv preprint arXiv:1810.04805}.

\bibitem[{Ghazvininejad et~al.(2016)Ghazvininejad, Shi, Choi, and
  Knight}]{ghazvininejad2016generating}
Ghazvininejad, M.; Shi, X.; Choi, Y.; and Knight, K. 2016.
\newblock Generating topical poetry.
\newblock In \emph{Proceedings of the 2016 Conference on Empirical Methods in
  Natural Language Processing}, 1183--1191.

\bibitem[{Goodfellow et~al.(2020)Goodfellow, Pouget-Abadie, Mirza, Xu,
  Warde-Farley, Ozair, Courville, and Bengio}]{goodfellow2020generative}
Goodfellow, I.; Pouget-Abadie, J.; Mirza, M.; Xu, B.; Warde-Farley, D.; Ozair,
  S.; Courville, A.; and Bengio, Y. 2020.
\newblock Generative adversarial networks.
\newblock \emph{Communications of the ACM}, 63(11): 139--144.

\bibitem[{Hu et~al.(2021)Hu, Shen, Wallis, Allen-Zhu, Li, Wang, Wang, and
  Chen}]{hu2021lora}
Hu, E.~J.; Shen, Y.; Wallis, P.; Allen-Zhu, Z.; Li, Y.; Wang, S.; Wang, L.; and
  Chen, W. 2021.
\newblock Lora: Low-rank adaptation of large language models.
\newblock \emph{arXiv preprint arXiv:2106.09685}.

\bibitem[{Hu et~al.(2017)Hu, Yang, Liang, Salakhutdinov, and
  Xing}]{hu2017toward}
Hu, Z.; Yang, Z.; Liang, X.; Salakhutdinov, R.; and Xing, E.~P. 2017.
\newblock Toward controlled generation of text.
\newblock In \emph{International conference on machine learning}, 1587--1596.
  PMLR.

\bibitem[{Kingma and Welling(2013)}]{kingma2013auto}
Kingma, D.~P.; and Welling, M. 2013.
\newblock Auto-encoding variational bayes.
\newblock \emph{arXiv preprint arXiv:1312.6114}.

\bibitem[{Lau et~al.(2018)Lau, Cohn, Baldwin, Brooke, and
  Hammond}]{lau2018deep}
Lau, J.~H.; Cohn, T.; Baldwin, T.; Brooke, J.; and Hammond, A. 2018.
\newblock Deep-speare: A joint neural model of poetic language, meter and
  rhyme.
\newblock \emph{arXiv preprint arXiv:1807.03491}.

\bibitem[{Li et~al.(2015)Li, Galley, Brockett, Gao, and
  Dolan}]{li2015diversity}
Li, J.; Galley, M.; Brockett, C.; Gao, J.; and Dolan, B. 2015.
\newblock A diversity-promoting objective function for neural conversation
  models.
\newblock \emph{arXiv preprint arXiv:1510.03055}.

\bibitem[{Li et~al.(2020)Li, Zhang, Liu, and Shi}]{li2020rigid}
Li, P.; Zhang, H.; Liu, X.; and Shi, S. 2020.
\newblock Rigid formats controlled text generation.
\newblock In \emph{Proceedings of the 58th annual meeting of the association
  for computational linguistics}, 742--751.

\bibitem[{Li et~al.(2022)Li, Thickstun, Gulrajani, Liang, and
  Hashimoto}]{li2022diffusion}
Li, X.~L.; Thickstun, J.; Gulrajani, I.; Liang, P.; and Hashimoto, T.~B. 2022.
\newblock Diffusion-LM Improves Controllable Text Generation.
\newblock \emph{arXiv preprint arXiv:2205.14217}.

\bibitem[{Lin(2004)}]{lin2004rouge}
Lin, C.-Y. 2004.
\newblock Rouge: A package for automatic evaluation of summaries.
\newblock In \emph{Text summarization branches out}, 74--81.

\bibitem[{Lin et~al.(2017)Lin, Li, He, Zhang, and Sun}]{lin2017adversarial}
Lin, K.; Li, D.; He, X.; Zhang, Z.; and Sun, M.-T. 2017.
\newblock Adversarial ranking for language generation.
\newblock \emph{Advances in neural information processing systems}, 30.

\bibitem[{OpenAI(2021)}]{openai2021gpt35turbo}
OpenAI. 2021.
\newblock GPT-3.5-turbo: Language Model.
\newblock \url{https://platform.openai.com/docs/guides/chatgpt}.

\bibitem[{Papineni et~al.(2002)Papineni, Roukos, Ward, and
  Zhu}]{papineni2002bleu}
Papineni, K.; Roukos, S.; Ward, T.; and Zhu, W.-J. 2002.
\newblock Bleu: a method for automatic evaluation of machine translation.
\newblock In \emph{Proceedings of the 40th annual meeting of the Association
  for Computational Linguistics}, 311--318.

\bibitem[{Pu and Demberg(2023)}]{pu2023chatgpt}
Pu, D.; and Demberg, V. 2023.
\newblock ChatGPT vs Human-authored Text: Insights into Controllable Text
  Summarization and Sentence Style Transfer.
\newblock \emph{arXiv preprint arXiv:2306.07799}.

\bibitem[{Sheng et~al.(2021)Sheng, Song, Tan, Ren, Ye, Zhang, and
  Qin}]{sheng2021songmass}
Sheng, Z.; Song, K.; Tan, X.; Ren, Y.; Ye, W.; Zhang, S.; and Qin, T. 2021.
\newblock Songmass: Automatic song writing with pre-training and alignment
  constraint.
\newblock In \emph{Proceedings of the AAAI Conference on Artificial
  Intelligence}, volume~35, 13798--13805.

\bibitem[{Sohl-Dickstein et~al.(2015)Sohl-Dickstein, Weiss, Maheswaranathan,
  and Ganguli}]{sohl2015deep}
Sohl-Dickstein, J.; Weiss, E.; Maheswaranathan, N.; and Ganguli, S. 2015.
\newblock Deep unsupervised learning using nonequilibrium thermodynamics.
\newblock In \emph{International Conference on Machine Learning}, 2256--2265.
  PMLR.

\bibitem[{Song, Meng, and Ermon(2020)}]{song2020denoising}
Song, J.; Meng, C.; and Ermon, S. 2020.
\newblock Denoising diffusion implicit models.
\newblock \emph{arXiv preprint arXiv:2010.02502}.

\bibitem[{Sutskever, Vinyals, and Le(2014)}]{sutskever2014sequence}
Sutskever, I.; Vinyals, O.; and Le, Q.~V. 2014.
\newblock Sequence to sequence learning with neural networks.
\newblock \emph{Advances in neural information processing systems}, 27.

\bibitem[{Tian and Peng(2022)}]{tian2022zero}
Tian, Y.; and Peng, N. 2022.
\newblock Zero-shot Sonnet Generation with Discourse-level Planning and
  Aesthetics Features.
\newblock \emph{arXiv preprint arXiv:2205.01821}.

\bibitem[{Touvron et~al.(2023)Touvron, Martin, Stone, Albert, Almahairi,
  Babaei, Bashlykov, Batra, Bhargava, Bhosale et~al.}]{touvron2023llama}
Touvron, H.; Martin, L.; Stone, K.; Albert, P.; Almahairi, A.; Babaei, Y.;
  Bashlykov, N.; Batra, S.; Bhargava, P.; Bhosale, S.; et~al. 2023.
\newblock Llama 2: Open foundation and fine-tuned chat models.
\newblock \emph{arXiv preprint arXiv:2307.09288}.

\bibitem[{Van~de Cruys(2020)}]{van2020automatic}
Van~de Cruys, T. 2020.
\newblock Automatic poetry generation from prosaic text.
\newblock In \emph{Proceedings of the 58th annual meeting of the association
  for computational linguistics}, 2471--2480.

\bibitem[{Yi et~al.(2020)Yi, Li, Yang, Li, and Sun}]{yi2020mixpoet}
Yi, X.; Li, R.; Yang, C.; Li, W.; and Sun, M. 2020.
\newblock Mixpoet: Diverse poetry generation via learning controllable mixed
  latent space.
\newblock In \emph{Proceedings of the AAAI conference on artificial
  intelligence}, volume~34, 9450--9457.

\bibitem[{Yu et~al.(2017)Yu, Zhang, Wang, and Yu}]{yu2017seqgan}
Yu, L.; Zhang, W.; Wang, J.; and Yu, Y. 2017.
\newblock Seqgan: Sequence generative adversarial nets with policy gradient.
\newblock In \emph{Proceedings of the AAAI conference on artificial
  intelligence}, volume~31.

\bibitem[{Zhang, Liu, and Zhang(2023)}]{zhang2023summit}
Zhang, H.; Liu, X.; and Zhang, J. 2023.
\newblock SummIt: Iterative Text Summarization via ChatGPT.
\newblock \emph{arXiv preprint arXiv:2305.14835}.

\bibitem[{Zhang and Song(2022)}]{zhang2022discup}
Zhang, H.; and Song, D. 2022.
\newblock DisCup: Discriminator cooperative unlikelihood prompt-tuning for
  controllable text generation.
\newblock \emph{arXiv preprint arXiv:2210.09551}.

\bibitem[{Zhang et~al.(2019)Zhang, Liu, Chen, Hu, Xu, and
  Mao}]{zhang2019generating}
Zhang, R.; Liu, X.; Chen, X.; Hu, Z.; Xu, Z.; and Mao, Y. 2019.
\newblock Generating Chinese CI with designated metrical structure.
\newblock In \emph{Proceedings of the AAAI Conference on Artificial
  Intelligence}, volume~33, 7459--7467.

\bibitem[{Zhang and Lapata(2014)}]{zhang2014chinese}
Zhang, X.; and Lapata, M. 2014.
\newblock Chinese poetry generation with recurrent neural networks.
\newblock In \emph{Proceedings of the 2014 Conference on Empirical Methods in
  Natural Language Processing (EMNLP)}, 670--680.

\bibitem[{Zhou et~al.(2023)Zhou, Jiang, Wilcox, Cotterell, and
  Sachan}]{zhou2023controlled}
Zhou, W.; Jiang, Y.~E.; Wilcox, E.; Cotterell, R.; and Sachan, M. 2023.
\newblock Controlled text generation with natural language instructions.
\newblock \emph{arXiv preprint arXiv:2304.14293}.

\end{thebibliography}

\end{document}